\documentclass{article}

\newcommand{\comment}[1]{}

\usepackage{amssymb}  % \checkmark
\usepackage{color}
\definecolor{orange}{RGB}{255,127,0}
\definecolor{brown}{RGB}{150,70,0}
\definecolor{green}{RGB}{127,255,127}
\definecolor{darkgreen}{RGB}{0,127,0}
\definecolor{blue}{RGB}{127,127,255}
\definecolor{lightblue}{RGB}{150,150,255}
\definecolor{darkblue}{RGB}{0,0,127}
\definecolor{red}{RGB}{255,90,90}
\definecolor{grey}{RGB}{127,127,127}
\definecolor{pink}{RGB}{255,180,180}

\usepackage{graphicx}

\title{A short note on estimating intelligence from user profiles in the context of universal psychometrics: \\ prospects and caveats
}
\author
{
	Jos\'{e} Hern\'{a}ndez-Orallo\\
	{\normalsize\em DSIC, Universitat Polit\`ecnica de Val\`encia, Spain}\\
	{\normalsize \tt jorallo@dsic.upv.es}
}
\date{}

\begin{document}

\maketitle

{
\abstract There has been an increasing interest in inferring some personality traits from users and players in social networks and games, respectively. This goes beyond classical sentiment analysis, and also much further than customer profiling. The purpose here is to have a characterisation of users in terms of personality traits, such as openness, conscientiousness, extraversion, agreeableness, and neuroticism. While this is an incipient area of research, we ask the question of whether cognitive abilities, and intelligence in particular, are also measurable from user profiles. However, we pose the question as broadly as possible in terms of subjects, in the context of universal psychometrics, including humans, machines and hybrids. Namely, in this paper we analyse the following question: is it possible to measure the intelligence of humans and (non-human) bots in a social network or a game just from their user profiles, i.e., by observation, without the use of interactive tests, such as IQ tests, the Turing test or other more principled machine intelligence tests? 
}

%%%%%%%%%%%%%%%%%%%%%%%%%%%%%%%%%%%%%%%%%%%%%%%%%%%%%%%%%%%%%%%%%%
%%%%%%%%%%%%%%%%%%%%%%%%%%%%%%%%%%%%%%%%%%%%%%%%%%%%%%%%%%%%%%%%%%
\section{Introduction}
%%%%%%%%%%%%%%%%%%%%%%%%%%%%%%%%%%%%%%%%%%%%%%%%%%%%%%%%%%%%%%%%%%
%%%%%%%%%%%%%%%%%%%%%%%%%%%%%%%%%%%%%%%%%%%%%%%%%%%%%%%%%%%%%%%%%%

Virtual environments created by games, social networks and virtual worlds are different from natural environments in many ways. One remarkably property is that there is freedom to detach the physical features of participants from the virtual appearance. This means that it is relatively easy to impersonate anyone and use anonymity. As a result, humans and bots can interact and participate in these environments in similar conditions. In addition, virtual environments, unlike real environments, usually record a great part of the things that happen in the environment, as actions, conversations, etc. This means that (almost) everything is recorded and can eventually tracked, processed and analysed afterwards. Consequently, virtual environments are an excellent playground to analyse and compare the behaviour of humans and machines.

User profiles are usually derived from the information that is constantly recorded and archived, in terms of what the users do, say, buy, etc. This inference of user interests, intentions, characteristics, preferences, and ultimately behaviours, is usually known as `user profiling'.
Of course, user profiling has been common in many other areas ---most especially when the user is a customer---, much before virtual environments were common like today. However, virtual environments record such an amount of information (as users `spend' so much time in them, or just connected through their gadgets) that it is now possible to build very rich (and accurate) user profiles. 
Profiles usually integrate given information (age, location, gender, etc.) with the inferred information (user preferences, activities, ...). 
Some {\em abilities} have also (occasionally) been part of the user profile, ``both mental and physical" \cite{golemati2007creating}. Very recently, some works have started to predict personality traits or abilities \cite{quercia2011our,golbeck2011predicting,bachrach2012personality,staiano2012friends,ortigosa2013predicting,kosinski2013private}. These works have focussed on humans ---more precisely, they have not questioned the humanness of the profiles. 

In this paper we analyse whether cognitive abilities in general, and intelligence in particular are measurable when the users may be humans or bots. In other words, and perhaps more interestingly, how easy is to fool these predictions when using machines instead of humans? Can we design a bot that learns to create profiles such that their predicted cognitive abilities are placed as desired? Can humans use these bots to look what they are not, e.g., more or less intelligent?

The rest of the paper is organised as follows. 
Section \ref{sec:upsycho} discusses how machine intelligence has been evaluated and briefly revisits the approach to evaluate human and machine intelligence in an integrated, non-anthropocentric way.
Section \ref{sec:noprofiles} reviews ways of inferring intelligence (or, rather, humanness) by observation, but without using profiles.
Section \ref{sec:profiles} discusses the central issue in this paper, how intelligence can be inferred from user profiles in humans and whether this is extensible to machines. 
Finally, section \ref{sec:discussion} finds several caveats about this and the ultimate interest of the possibility of inferring intelligence in this way.

%%%%%%%%%%%%%%%%%%%%%%%%%%%%%%%%%%%%%%%%%%%%%%%%%%%%%%%%%%%%%%%%%%
%%%%%%%%%%%%%%%%%%%%%%%%%%%%%%%%%%%%%%%%%%%%%%%%%%%%%%%%%%%%%%%%%%
\section{Measuring cognitive abilities of humans and machines: universal psychometrics} \label{sec:upsycho}
%%%%%%%%%%%%%%%%%%%%%%%%%%%%%%%%%%%%%%%%%%%%%%%%%%%%%%%%%%%%%%%%%%
%%%%%%%%%%%%%%%%%%%%%%%%%%%%%%%%%%%%%%%%%%%%%%%%%%%%%%%%%%%%%%%%%%

The evaluation of the cognitive abilities of biological beings has been been addressed by psychometrics \cite{Sternberg2000,Boorsboom2005}, for humans, and by cognitive science \cite{Shettleworth2010} and comparative psychology, when animals are also considered. 
When talking about machines, things are much more scattered and immature: specific tests (Turing Test \cite{turing1950,oppydowe2011}, the total Turing Test \cite{schweizer1998truly} and other sensorimotor variations \cite{Neumann-etal09}, the Bot Prize \cite{hingston2010new}, CAPTCHAs \cite{von2004telling}, the use of human IQ tests for machines \cite{Evans1963,evans1964,Evans1965,iq,BringsjordSchimanski2003,bringsjord2011} , the machine intelligence quotient \cite{zadeh1976miq,bien2002machine}, ... ), competitions (RL competition \cite{whiteson2010reinforcement}, Robocup \cite{kitano1997robocup}, general game playing competition \cite{genesereth2005general}, planning competition \cite{Planningcompetition}, ...) and landmarks (Deep blue  \cite{deepblue2002}, Watson \cite{ferrucci2010building}). 

The problems of the Turing Test and other non-systematic ways of evaluating machine intelligence have been spotted by many (see, e.g., \cite{HernandezOrallo2000a}). Similarly, the justification of why IQ tests are not for machines (or for biological systems other than humans) is more cumbersome \cite{IQnotformachines}. The inadequacy of these approaches has led to a more principled, non-anthropocentric approach of evaluating machine intelligence, usually based on concepts such algorithmic information theory, Solomonoff prediction, the MML principle, compression, etc. we refer to the following references for a comprehensive picture: \cite{Dowe-Hajek1997a,Dowe-Hajek1997b,HernandezOrallo-MinayaCollado1998,Dowe-Hajek1998,mahoney1999text,HernandezOrallo2000a,HernandezOrallo00b,HernandezOrallo00c,HernandezOrallo00d,Legg-Hutter2007,Hibbard2009,HernandezOrallo-Dowe2010,HernandezOrallo10b,AGI2011Evaluating,AGI2011Compression,AGI2011DarwinWallace,CAEPIA2011,LeggVeness2011,Manchester2012,AISB-AICAP2012a,AISB-AICAP2012b,AGI2012social,potential,hernandez2013potential,hernandez2013complexity,hernandez2013universality}.

There is a notion that has emerged from the previous works: the idea of a `universal (intelligence) test', first illustrated in \cite{HernandezOrallo-Dowe2010}, as a test that could be administered to any kind of subject, of any speed, and interrupted anytime. The notion of a `test for all' is appealing and soon generated widespread interest \cite{TheEconomist2011,NewScientist2011}. The extent and breadth of universal tests has been analysed in \cite{hernandez2013universality}. This study suggests that there are many challenges to be solved if we really want a most general test. However, the notion of evaluating humans, animals and machines with the same principles has led to a new discipline called {\em universal psychometrics} \cite{upsychometrics,upsychometrics2}, defined as {\em the general evaluation of cognitive abilities of any kind of subject (with or without universal tests)}.

One of the motivations for universal psychometrics has precisely been the existence of a plethora of bots, agents, robots, hybrids and collectives thereof and other kinds of artificial agents in social networks, games, messaging networks, virtual worlds and other Internet-based spaces. This increasing variety of systems in an increasing variety of environments suggests more emphatically that the CAPTCHA approach or the use of IQ tests is too anthropomorphic and limited to determine intelligence appropriately.

%%%%%%%%%%%%%%%%%%%%%%%%%%%%%%%%%%%%%%%%%%%%%%%%%%%%%%%%%%%%%%%%%%
%%%%%%%%%%%%%%%%%%%%%%%%%%%%%%%%%%%%%%%%%%%%%%%%%%%%%%%%%%%%%%%%%%
\section{Measuring intelligence without profiles} \label{sec:noprofiles}
%%%%%%%%%%%%%%%%%%%%%%%%%%%%%%%%%%%%%%%%%%%%%%%%%%%%%%%%%%%%%%%%%%
%%%%%%%%%%%%%%%%%%%%%%%%%%%%%%%%%%%%%%%%%%%%%%%%%%%%%%%%%%%%%%%%%%

We will now focus on virtual environments and see the specific alternatives here. The best way to evaluate the intelligence of an agent (be it a human or a machine) in a virtual environment is to use an intelligence test (but not a CAPTCHA or a Turing Test variant, as discussed above). Finding good tests for this is the goal of universal psychometrics. Nonetheless, the question we ask in this paper is whether we can evaluate intelligence from the behaviour of the agents, without actually using tests on purpose.
So we will review and investigate how intelligence can be measured observationally without profiles (below) and with profiles (in the next section).

We can explore whether there are other ways of (loosely) interacting with or observing agents in order to get estimations of their intelligence (or related traits). For instance, in games there is an increasing interest in the concept of {\em believability} \cite{hingston2012believable}. Several techniques are used to determine whether a bot behaves like a human. This ends up falling back into the area of Turing Tests for bots \cite{hingston2009turing,hingston2010new} such as BotPrize \cite{hingston20092k}, although it is not always suggests as a test. 
As a result, this approach has the same benefits and inconveniences of the use of a Turing Test in other environments (e.g., the Loebner Prize). On one hand, we can use intelligent judges to implement the test, and there is flexibility on the issues or patterns the judges can look at. On the other, these tests are getting increasingly easier to crack, the behaviour of the evaluee may change because of the realisation of being examined, they are not automated (as human judges are needed) and, more importantly, it is humanness what is being measured instead of intelligence. 

Similarly, we could think of CAPTCHAs \cite{von2004telling} as an automated solution, but they are still more anthropocentric than the Turing Test and evaluate some human abilities which are poorly related with intelligence (such as reading distorted text). Also, it is again a disruptive approach (if these tests are executed regularly).

Alternatively, the principled approach represented by universal psychometrics may be useful in games, since players can be evaluated with rewards, without actually looking as a test. The problem is that intelligence is said to correlate with performance on a {\em variety} of games (see, e.g., \cite[sec. 6.5]{HernandezOrallo-Dowe2010}), not only one specific game. Also, in social networks, there is no clear notion of reward, as people are just here for fun or for socialising, not for getting rewards.

From here, it seems difficult (if not impossible) to evaluate intelligence without actually alienating the agent from its context and administering an intelligence test on purpose. Is there any alternative? This is what we want to explore in this paper, the possibility of estimating intelligence by examining user's profiles and logs. This is what we do next.

%%%%%%%%%%%%%%%%%%%%%%%%%%%%%%%%%%%%%%%%%%%%%%%%%%%%%%%%%%%%%%%%%%
%%%%%%%%%%%%%%%%%%%%%%%%%%%%%%%%%%%%%%%%%%%%%%%%%%%%%%%%%%%%%%%%%%
\section{Measuring intelligence with profiles} \label{sec:profiles}
%%%%%%%%%%%%%%%%%%%%%%%%%%%%%%%%%%%%%%%%%%%%%%%%%%%%%%%%%%%%%%%%%%
%%%%%%%%%%%%%%%%%%%%%%%%%%%%%%%%%%%%%%%%%%%%%%%%%%%%%%%%%%%%%%%%%%

A person can be judged by what she is able to do or what she has done. Intelligence is usually associated to what she is able to do. Nonetheless, we frequently infer abilities by looking at what has been done. In the context of virtual environments, an agent creates a history, a log or, if properly arranged, generalised and integrated by some user-provided information, a profile. From a user profile, in a social network, a game or a virtual world, we can know what the user does, how she talks, what she says, what friends she has, what things she like, what she buys, etc. As said in the introduction, there is a strong interest in extracting value from all this information in order to customise applications, products and interfaces to this behaviour.

One kind of information that has been recently extracted from user profiles is user personality, decomposed into several traits, such as openness, conscientiousness, extraversion, agreeableness, and neuroticism\footnote{The traits may vary depending on the model: the P.E.N. model, the Big Five or the Alternative Five model.} \cite{quercia2011our,golbeck2011predicting,bachrach2012personality,ortigosa2013predicting}. These traits are extracted from what the user does, but they can also be inferred in a more networked (or social) way, by using information about the user's friends \cite{staiano2012friends}. In fact, this turns out to be a transductive problem (instead of an inductive problem), as the correct categorisation of some users may be useful to categorise her friends afterwards, who may be more difficult to categorise on a first attempt.

Most interestingly, one recent work has included a very diverse set of traits, including some ideological and personal ones, such as gender, political views, drug addiction, sex orientations, etc., along with the above mentioned traits, but also including {\em intelligence} \cite{kosinski2013private}. This work shows that it is possible to infer the user's intelligence from her profiles, using a very simple linear model after singular-value decomposition, yielding to a prediction with 0.39 of Pearson correlation. This work highlights the high predictability of the ``Likes'' attribute: ``for example, the best predictors of high intelligence include `Thunderstorms,' `The Colbert Report,' `Science,' and `Curly Fries,' whereas low intelligence was indicated by `Sephora,' `I Love Being A Mom,' `Harley Davidson,' and `Lady Antebellum'.'' While some of them may have some sense, such as `Thunderstorms' or `Science', others may have a more difficult explanation, such as `Curly Fries'. Independently from whether the model is understandable or not, it is a fact that we can infer positively-correlated information about {\em human} user intelligence with a few profile data. This possibility unveils many possible applications. In personnel selection, we can use this information to eliminate some recruit tests during the selection process. In teaching and other academic areas, this information can be used to better choose or customise assignments to the particular student's abilities. While the accuracy may be worse than on-purpose intelligence test, the good thing about this approach is that it goes smoothly and inadvertently for the subjects.

However, what happens when we consider non-human users as well? All the previous studies do not consider this possibility. 
Certainly, if we use bots, many good predictors will be less so. This is especially the case with current technology, as any bot that is able to manage in a social network, a game or a virtual world, will probably have a very {\em special} profile. This will be more or less so depending on the variety of actions and communication that the environment can offer. If natural language conversation exists in a free way (not with a restricted set of phrases, as in some massive multiplayer online games for children) % Club Penguin
then a much more complex analysis in terms of natural language processing could reveal much more information that a mere ``likes'' analysis (or other simple attributes) but will also be much more difficult. Also, in games, the actions, strategies, and alliances may reveal part of the users' intelligence. In addition, the assessment or comments received by other users can also be a good source of information. On occasions this could get closer to an informal Turing Test, if some other users make (intentional or unintentional) judgements about the user's level of intelligence in their comments. Finally, in those cases where users have some goals (e.g., games) or are asked to do something (working teams), we could try to evaluate their performance or accomplishment in these tasks, or at least what kinds of subgoals or subtasks the agent is able to accept and those that the user rejects, in order to estimate their cognitive abilities.

Overall, there is a huge space of exploration in order to determine the information about user actions and conversation that may useful for estimating intelligence. The attributes and information required may vary depending on the technology and the expected degree of bot intelligence, the existence of hybrids, etc. Of course, in any case, we would require some supervised data, i.e., some of the users (humans and bots) should be properly evaluated before, in order to take them as a reference. While the evaluation of human intelligence is reliable, the evaluation of machine intelligence is still in its infancy. 

The previous picture shows a highly challenging problem. If the problem can be eventually be solved by approximating (with some degree of reliability) some cognitive abilities, then there may be some interesting applications when thinking about humans and bots together. 
The most likely area of application is when hybrid groups or teams must be created to accomplish a task: a more appropriate selection of humans and bots can be performed if we have an estimation of their cognitive abilities. Similarly, in games, we can use this information to create teams such that the match or competition is more equilibrated. On the other hand, if the problem cannot be solved in this way with any minimal degree of reliability, this may also useful as a source of valuable and interesting information for understanding why some predictors may only work for humans while others might be more universal. This also suggests that it may be more informative to use predictive techniques that yield comprehensible models, such as decision trees or that indirectly convert model into comprehensible ones \cite{pazzani1997beyond,domingos1998knowledge,comprehensible2002,mimetic2003,van2007seeing}.

%%%%%%%%%%%%%%%%%%%%%%%%%%%%%%%%%%%%%%%%%%%%%%%%%%%%%%%%%%%%%%%%%%
%%%%%%%%%%%%%%%%%%%%%%%%%%%%%%%%%%%%%%%%%%%%%%%%%%%%%%%%%%%%%%%%%%
\section{Discussion} \label{sec:discussion}
%%%%%%%%%%%%%%%%%%%%%%%%%%%%%%%%%%%%%%%%%%%%%%%%%%%%%%%%%%%%%%%%%%
%%%%%%%%%%%%%%%%%%%%%%%%%%%%%%%%%%%%%%%%%%%%%%%%%%%%%%%%%%%%%%%%%%

We have just explored the general question that gives name to this paper. The possibility is challenging but may have interesting benefits if ever successful. However, at this point, we have failed to make an assessment of its viability in the short or long terms. In fact, we should not try to do this assessment without considering some further difficulties. Next we comment on several caveats about the approach of inferring intelligence using user profiles.

The first important caveat comes from the type of techniques that are usually employed for bots {\em believability} is desired. For instance, some bots are based on ``human traces'', such as \cite{ karpov2012believable}. These systems try to replicate ``chunks'' of human behaviour. Locally, these chunks or traces may look consistent and, if properly put together, may also look consistent globally as well. Not surprisingly, this approach is a generalisation of how the first chatterbots worked, as Weizenbaum's Eliza \cite{weizenbaum1966eliza}. The problem is that, at least initially, we expect that the analysis of profiles is based on statistical criteria rather than semantic ones. With this kind of analysis, the profile may be just an aggregation of the chunk's profiles, which may correspond to intelligent profiles (as extracted from intelligent humans), while the bot clearly has no intelligence at all.

Another general problem (which is not only particular to machines) is that users may be playing a role. This means that they may try to look more intelligent than they really are (by, e.g., adding `science' to their `likes'). It can also be the other way round, as many may try to look less intelligent than they really are to give a more sociable and accessible impression. Closely related, we also have the problem of impersonation, where the role played is to feign, and not necessarily focussing on the cognitive abilities but rather on the personality.
% For bots, many of these problems as well. Aslo, we have the notion of character or avatar... actors and personalities, and it is more related to (artistic) performance. 

Nonetheless, one of the most important caveats is when bots are not designed to perform a given task or offering a service, but they are (also) conceived to get good results in these intelligence estimations, i.e., to fool the estimations. A good way of doing this is by looking at how the methods that infer intelligence from user domains could work and learn from them. This `adversarial learning' scenario would probably be endless, and resembles what is happening today with CAPTCHAs, which need to be replaced as some bots are able to pass them. 

Another (related) problem of measuring in an uncontrolled setting is that bots can use human computation. For instance, they could just ask what to do or what to say to some human users. In fact, if they are able to verbalise their objectives, they could exploit their network of friends to achieve their goals, and look more intelligent. This issue also happens with CAPTCHAs, as some malicious bots forward their CAPTCHAs to some specialised sites where they are being passed by humans.

Conversely, humans could use machines to look more intelligent. This is already happening in many contexts, as humans are currently enhanced by calculators, smart phones, web search engines, encyclopaedias, translators, thesaurus, puzzle solvers, etc. In a more sophisticated way, humans could use machines to create any fake profile, by letting a bot do part of their actions and even conversation. Bots are already used on some social networks to pretend that the user is awake, or working, or doing some activity. The possibilities of hybridisation (with either humans and bot taken the dominant role) are endless.

%Finally, there are ethical issues as well, as inferring intelligence from humans and bots 
% For humans... Also ethical issues: some companies may be interested in quickly assessing customer’s intelligence in order to keep the silliest customers.

Despite all these caveats, the possibility of inferring intelligence or other cognitive abilities from behaviour in a virtual environment deserves some attention. Many bots are starting to be able to solve some specific tasks derived from complex problems, where groups of people and artificial agents collaborate. Being able to infer the intelligence (and other cognitive abilities) of all the partners in a team or group may be very useful in order to allocate the most appropriate tasks.

In more general terms, the interest in intelligence estimation from profiles will probably be increased as bot intelligence gets closer and closer to (and eventually beyond) human intelligence. For the moment, I think that this may be a secondary (but interesting) scenario to better determine what universal psychometrics can be and possibly refine the notion of universal test.

%%%%%%%%%%%%%%%%%%%%%%%%%%%%%%%%%%%%%%%%%%%%%%%%%%%%%%%%%%%%%%%%%%
%%%%%%%%%%%%%%%%%%%%%%%%%%%%%%%%%%%%%%%%%%%%%%%%%%%%%%%%%%%%%%%%%%
%\section*{Acknowledgements}
%%%%%%%%%%%%%%%%%%%%%%%%%%%%%%%%%%%%%%%%%%%%%%%%%%%%%%%%%%%%%%%%%%
%%%%%%%%%%%%%%%%%%%%%%%%%%%%%%%%%%%%%%%%%%%%%%%%%%%%%%%%%%%%%%%%%%

\bibliography{biblio}

\end{document}